\documentclass[letterpaper, 10 pt, conference]{ieeeconf}  % Comment this line out if you need a4paper

\IEEEoverridecommandlockouts                              % This command is only needed if 
\usepackage{graphics} % for pdf, bitmapped graphics files
\usepackage{epsfig} % for postscript graphics files
\usepackage{mathtools}
\usepackage{times} % assumes new font selection scheme installed
\usepackage{amsmath} % assumes amsmath package installed
\usepackage{amssymb}  % assumes amsmath package installed
\usepackage{bm}
\usepackage{multirow}
\usepackage{makecell}
\usepackage{bbding}
\usepackage{bbm}

\usepackage[utf8]{inputenc} % allow utf-8 input
\usepackage[T1]{fontenc}    % use 8-bit T1 fonts
\usepackage{hyperref}       % hyperlinks
            
\usepackage{url}            % simple URL typesetting
\usepackage{booktabs}       % professional-quality tables
\usepackage{amsfonts}       % blackboard math symbols
\usepackage{nicefrac}       % compact symbols for 1/2, etc.
\usepackage{microtype}      % microtypography
\usepackage{xcolor}         % colors
\usepackage{cleveref}
\usepackage{physics}
\usepackage{hyperref}
\usepackage{graphicx}
\usepackage{float}
\usepackage{multirow}
\usepackage{diagbox}

\usepackage{algorithm}
\usepackage{algpseudocode}
\usepackage{arydshln}

\usepackage{multirow}
\usepackage{listings}
\usepackage{graphicx}
\usepackage{siunitx} % \si{\degree}
            
\definecolor{dkgreen}{rgb}{0,0.6,0}
\definecolor{gray}{rgb}{0.5,0.5,0.5}
\definecolor{mauve}{rgb}{0.58,0,0.82}
\title{\LARGE \bf
AdaFSNet: Time Series Classification Based on Convolutional Network with a Adaptive and Effective Kernel Size Configuration
}

\author{Haoxiao Wang, Bo Peng, Jianhua Zhang, Xu Cheng
\thanks{Haoxiao Wang, Bo Peng, Jianhua Zhang and Xu Cheng are with School of Computer Science and Engineering, Tianjin University of Technology}
\thanks{This work was supported by the National Natural Science Foundation of China Under Grant 62306212.}
\thanks{%\textsuperscript{\Letter} 
{\textdagger} Corresponding to Xu Cheng (xu.cheng@ieee.org) and Jianhua Zhang (zjh@email.tjut.edu.cn)}
}

\begin{document}
\maketitle
%===============================================================

\begin{abstract}
Time series classification is one of the most critical and challenging problems in data mining, existing widely in various fields and holding significant research importance. Despite extensive research and notable achievements with successful real-world applications, addressing the challenge of capturing the appropriate receptive field (RF) size from one-dimensional or multi-dimensional time series of varying lengths remains a persistent issue, which greatly impacts performance and varies considerably across different datasets.
In this paper, we propose an Adaptive and Effective Full-Scope Convolutional Neural Network (AdaFSNet) to enhance the accuracy of time series classification. This network includes two Dense Blocks. Particularly, it can dynamically choose a range of kernel sizes that effectively encompass the optimal RF size for various datasets by incorporating multiple prime numbers corresponding to the time series length.
We also design a TargetDrop block, which can reduce redundancy while extracting a more effective RF. To assess the effectiveness of the AdaFSNet network, comprehensive experiments were conducted using the UCR and UEA datasets, which include one-dimensional and multi-dimensional time series data, respectively. Our model surpassed baseline models in terms of classification accuracy, underscoring the AdaFSNet network's efficiency and effectiveness in handling time series classification tasks.
\end{abstract}

% \keywords{Object Grasping, Optical Flow, Transparent Objects} 

%===============================================================================

\section{Introduction}

%Robotic object grasping, a fundamental task in robotics, presents a multifaceted challenge due to the inherent diversity of objects encountered in real-world scenarios. Objects come in various shapes, sizes, materials, and textures, demanding versatile and adaptive grasping strategies. Many established approaches~\cite{xx, xx, xx} to object grasping have relied upon point-cloud cameras and extensive 3D information for grasp pose estimation. While effective in handling diverse object shapes, these methods may struggle to cope with transparent and specular objects that point-cloud cameras cannot reliably capture. 
Time series classification (TSC) is a crucial task in machine learning and data analysis, with the goal of predicting the category of a given time series \cite{yang200610, 10.1145/2379776.2379788}. This type of data analysis is prevalent across numerous fields, such as finance, environmental studies, and healthcare, demonstrating its wide-ranging applicability and importance \cite{hu2016classification, lai2023faithful, bagnall2017great}.
TSC research and applications have garnered significant interest in various domains. For instance, in environmental studies, TSC tasks are used to identify different types of vegetation by analyzing temporal optical signals from satellites \cite{RUWURM2020421}. In healthcare, TSC assists in distinguishing types of heartbeats from electrocardiogram data \cite{xie2021multi}. Additionally, in the field of animal behavior analysis \cite{lai2023multimodal}, TSC is employed to classify behaviors using data collected from Internet of Things devices. These examples highlight the diversity and significance of TSC tasks in multiple areas of study and application.

A significant difficulty in Time Series Classification (TSC) tasks lies in instructing models about the appropriate time scales for feature extraction. Traditional machine learning approaches have devoted substantial efforts to capturing critical time scales, but as the length of time series data increases, the computational resource requirements grow dramatically, posing a significant challenge.
The shapelet methods, as described by Hills \cite{hills2014classification}, utilize shapelets—subsequences within time series that exemplify a specific class—to gauge the similarity between various time series. This approach transforms the data based on these shapelets and then selects the optimal number of shapelets to avoid overfitting. The training complexity of the Shapelet Transform is proportional to the square of the number of training examples and the fourth power of the time series length, represented as $O(n^2l^4)$.
Other methods, as mentioned in \cite{schafer2015boss} and \cite{gorecki2013using}, assess the similarity of two time series by measuring their distance using a specific metric. However, these approaches require extensive research to find the crucial time scales. Similarly, recent advancements in deep learning methods have revealed a significant focus on addressing the issue of time scale. Generally, choosing the scales for feature extraction in 1D-CNN is considered a problem of hyper-parameter selection.
% MCNN \cite{cui2016multi} uses a grid search to find kernel sizes while also finding the best RF for a 1D-CNN. Tapnet \cite{zhang2020tapnet} additionally considers the dilation steps, and \cite{chen2021multi} also considers the number of layers. These aspects are vital for the RF in CNNs and their effectiveness in Time Series Classification.
Zhou \cite{ZHOU2016358} developed a technique using the Temporal Convolutional Network (TCN) to detect and locate unusual activities in crowded scenes in video sequences. Similarly, Razavian \cite{razavian2015temporal} applied the TCN for devising a diagnostic method from lab tests. However, there are ongoing challenges with the TCN's receptive field, particularly in fully leveraging historical data. As the number of hidden layers increases, so does the receptive field, leading to greater computational demands and complexity. This presents a significant hurdle in optimizing the TCN for efficient data utilization and processing.

To avoid those complicated and resource-consuming search efforts, inspired by OSCNN \cite{tang2020omni}, we propose the Adaptive Full-Scope Convolutional Neural Network
 (AdaFSNet), which utilizes the robust capability of the OS-Block to optimally determine the most effective Receptive Field and kernel size. Specifically, we design a unique TargetDrop layer to selectively identify the most effective channels for capturing temporal information while maintaining the relevant convolutional kernel. To enhance the model's accuracy, we have advanced our approach by integrating two dense blocks. These blocks utilize the convolution kernels preserved by the TargetDrop layer, thereby establishing an optimal Receptive Field. Furthermore, applying the AdaFSNet across various Time Series (TS) datasets is straightforward, as it involves selecting the largest prime number based on the length of the TS.

\begin{figure}[t]
\begin{center}
    \includegraphics[width=1.0\linewidth, height=5.0cm]{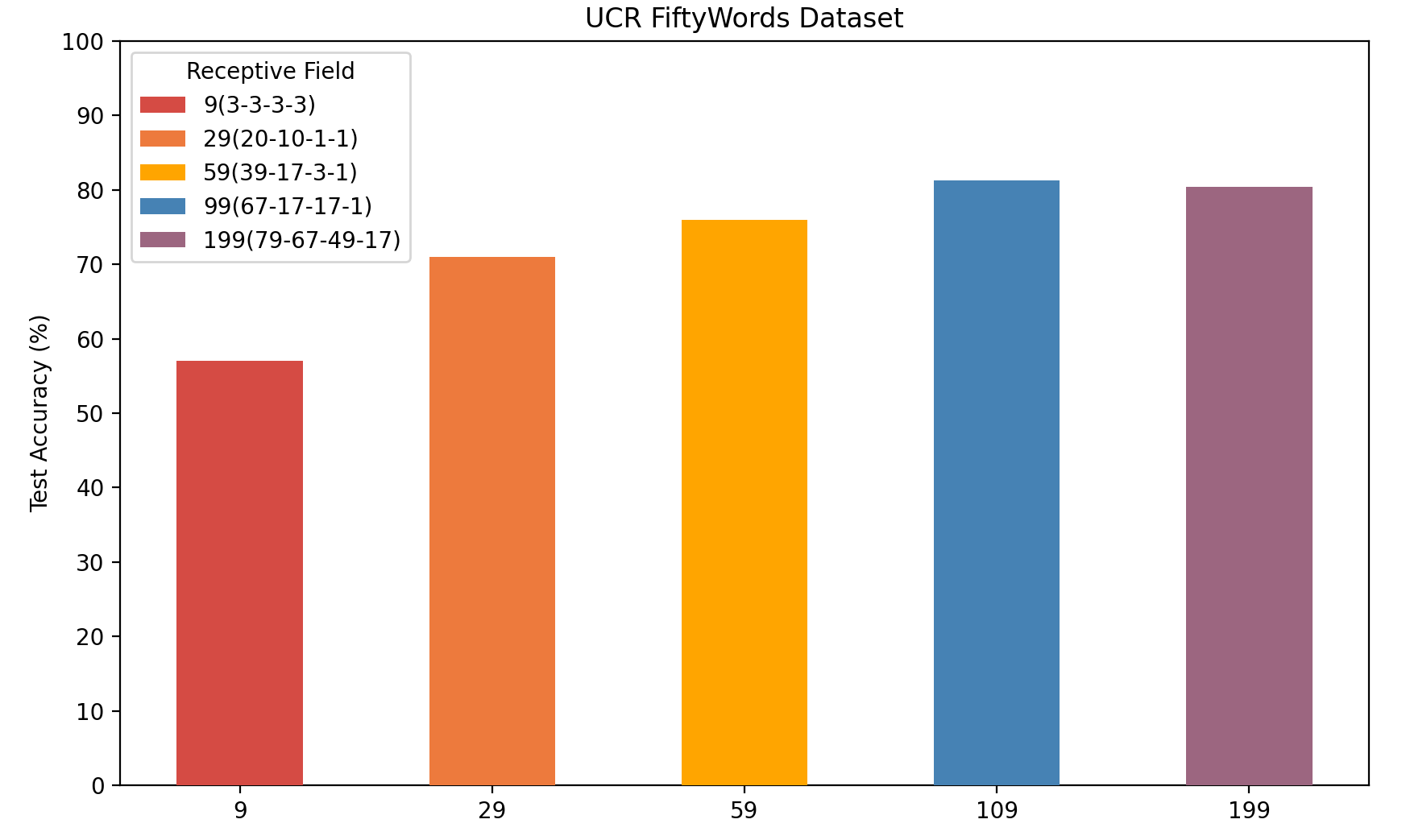}
\end{center}
\caption{\textbf{Analysis To RF Size.} We trained the model on the Fiftywords dataset in UCR, where the horizontal coordinate represents the RF size, and we identified the kernel configuration of each 1D-CNN.  For example, 9 (3-3-3-3) indicates that 1D-CNN has four layers and the receptive field size is 9, from the initial to the final layer, the kernel sizes are consistently set at 3 for each layer. As the receptive field size increases, there is a corresponding increase in accuracy. However, once a specific receptive field size is reached, further increasing it does not lead to improved accuracy; instead, it significantly escalates the computational demands of the model.}
\label{fig:teasor}
\vspace{-0.6cm}
\end{figure}

Our experiments demonstrate that this method achieves consistent state-of-the-art performance across some benchmarks and baselines in Time Series Classification.
The primary dataset utilized in our study was the UCR \& UEA dataset\cite{bagnall2018uea}, which encompasses numerous datasets across various domains, including  speech recognition, human activity recognition, spectrum analysis and healthcare.
Thorough experimentation conducted on these benchmark datasets reveals that our model adaptively captures time series features across multiple scales. Not only does it exhibit superior classification performance, but it also demonstrates swift training speeds and effortless convergence. To summarize, the contributions can be outlined as follows:
\begin{itemize}
    \item We present a novel Adaptive and Effective Full-Scope Convolutional Neural Network (AdaFSNet) to enhance the accuracy of time series classification, where the kernel sizes are decided by a simple and universal rule.
    \item To reduce the redundancy of too many paths to find the optimal RF size, we designed an attention module to adaptively select the best kernel size for good performance and to reduce the amount of calculation. At the same time, we added a new ResNet-like module behind the OS-Block that applies the best kernel size selected by the dropout layer to enhance the model's performance.
    \item We conducted experiments for both multi-view and sparse-view scenarios, Our model excels in generalizing for time series classification while maintaining a low computational resource requirement. It also seamlessly integrates with various network architectures, enhancing its adaptability.
\end{itemize}

\section{Related Work}
\subsection{Targeted dropout}  
The Dropout module \cite{gomez2019learning}, widely employed for the pruning and sparsification of neural networks, is particularly compatible with resource-limited devices. The Dropout model exhibits a multitude of variations, such as AttentionDrop \cite{ouyang2019attentiondrop} and DropPath \cite{zoph2018learning}. 
$L^0$ regularization, as introduced in \cite{louizos2017learning}, employs a modified version of concrete Dropout to apply regularization to the network's weight parameters, thereby controlling and inducing sparsity in the network. In their respective works, \cite{wang2017structured} and \cite{zhu2022targetdrop} put forward a Dropout-based pruning approach, involving a method to gradually fine-tune Dropout rates towards both zero and one. Our main focus in this research is the utilization of Attention Dropout to enhance the model's capability in capturing temporal information from time series data.

\subsection{Adaptive receptive field}
Adaptive receptive field techniques, as proposed in prior research\cite{han2018optimizing, wang2018deep}, demonstrate significant efficacy. These techniques dynamically adjust receptive field sizes during the training phase, mainly by applying a learned weight mask to the kernels.
In the ZoomOut\cite{ismail2020inceptiontime} method, four differently scaled inputs are used during the inference process. This approach is designed to effectively capture both contextual and local details.
In contrast, the AdaFSNet, another innovative approach, establishes connections between kernels of varying sizes to create diverse receptive fields, emphasizing critical dimensions. The adaptive receptive field excels in time series classification tasks, enabling 1D-CNNs to optimize their receptive field sizes. Its mathematical robustness, AdaFSNet, can be adapted for time series vision tasks, by employing prime size designs along the time dimension. This adaptation is crucial since similar regions of interest might span varying time scales in different datasets, and employing kernels of various sizes increases the likelihood of capturing these appropriate scales.

\subsection{Inception module}
Stacking CNN layers in deep convolutional neural networks is a common strategy to enhance performance, but it comes with two major drawbacks. Firstly, excessively deep networks are more susceptible to overfitting, particularly when dealing with smaller training datasets. Secondly, the computational resources required for training such networks experience a substantial increase, presenting challenges in terms of efficiency and resource management.
The Inception module was created to effectively pinpoint the most efficient sparse structure within a convolutional network, as described by Szegedy et al \cite{szegedy2017inception}. in their paper. This innovation significantly reduces the requisite number of parameters and the computational resources needed, addressing key issues in network design and efficiency.
Our AdaFSNet is inspired by this inception structure \cite{szegedy2017inception}.
Unlike these traditional methods, the AdaFSNet simplifies the process by eliminating the need for complex pre-training methods to weigh important scales. 
% For example, what is the maximum length to stop? How do they choose the depth of the neural network? How they choose The tolerance or ratio of their sequence?

\section{Method}

\begin{figure*}[t]
\begin{center}
    \includegraphics[width=0.99\linewidth, height=9.8cm]{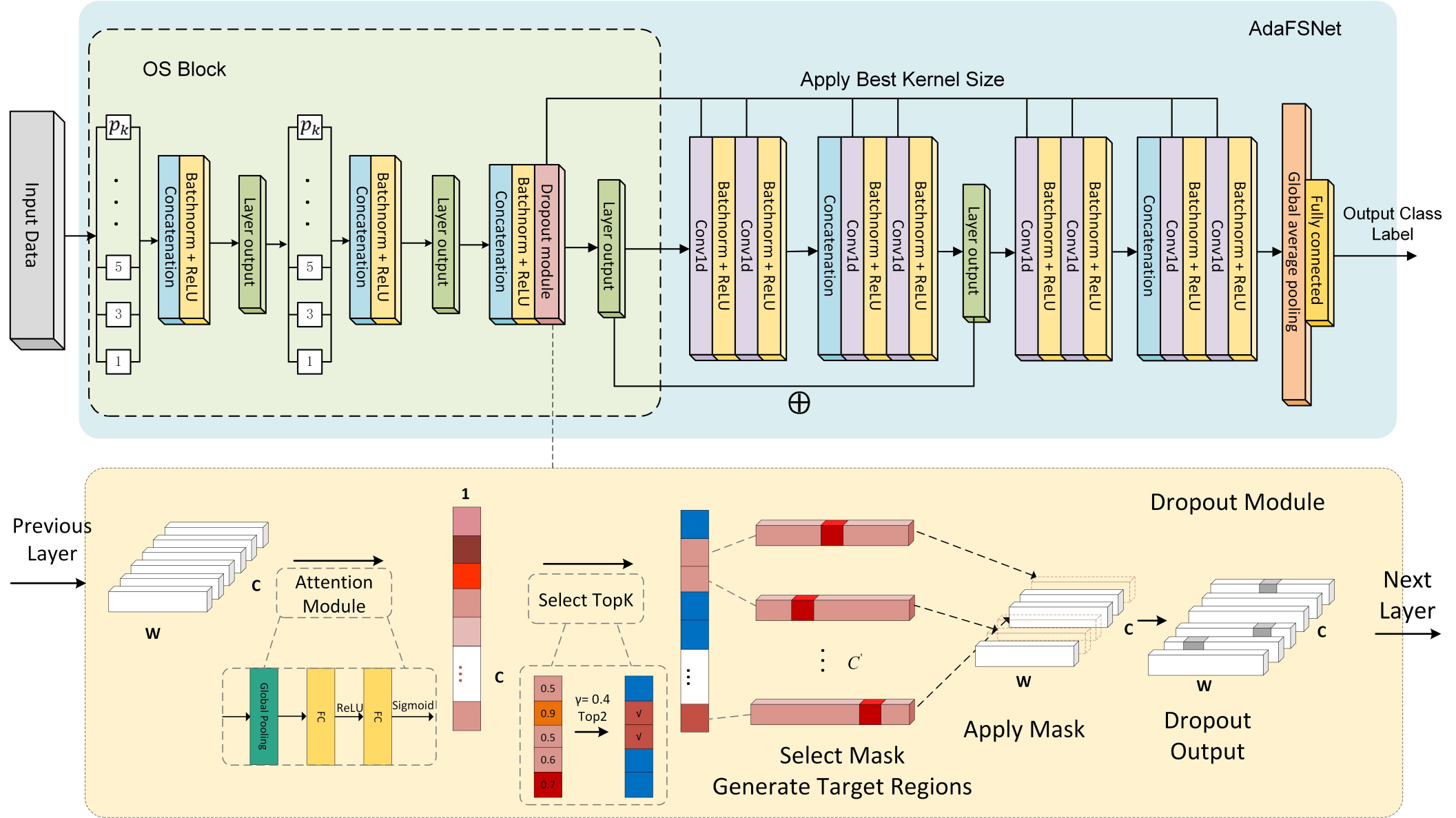}
\end{center}
% \vspace{-0.5cm}
\caption{\textbf{
% Demonstration for the pipeline of RGBGrasp.
AdaFSNet Pipeline.
} 
Goldbach's conjecture states that any even number can consist of two prime numbers.
Base on this theory,  with the OS-Block structure in the upper left of this image, which ensures the encompassment of all sizes of receptive fields. 
But not all prime numbers are suitable as kernel sizes. Therefore, we add a targeted dropout layer of attention module in the process to screen out more important RF, reduce redundancy and improve the accuracy of the model. 
In order to maximize the potential of PS-Block to cover all RF, We also add ResNet-like with 2Dense-Block module, whose kernel size is extracted from Dropout module. AdaFSNet achieves a series of SOTA performances on UCR \& UEA datasets and we list the details in Table II.
}
% \vspace{-0.7cm}
\label{fig:method}
\end{figure*}

\subsection{Problem Statement}
% Time Series data can be denoted as 
% $\textit{\textbf{X}} = [\textit{\textbf{x}}_1, \textit{\textbf{x}}_2, \ldots, \textit{\textbf{x}}_m]$, where $m$ is the number of variates. For univariate Time Series data, $m = 1$ and for $m > 1$, the TS are multivariate. Each variate is a vector of length $l$. A TS dataset, which has $n$ data and label paris, can be donoted as: $\mathbb{D} = \left\{(\textit{\textbf{X}}^1, y^1), (\textit{\textbf{X}}^2, y^2), \ldots, (\textit{\textbf{X}}^n, y^n)\right\}$, where $(\textit{\textbf{X}}^*, y^*)$ denotes the TS data $\textit{\textbf{X}}^*$ belongs to the class $y^*$. The task of Time Series Classification is train a classifier that can predict the class label $y^*$ when given a TS $\textit{\textbf{X}}^*$.
Let the time series dataset be denoted as $\mathcal{T}$, containing $N$ time series samples. Each sample is composed of time series data and a corresponding class label. Specifically, each sample can be represented as a pair $(\mathbf{S}, c)$, where $\mathbf{S}$ is the time series data and $c$ is its associated class label.
The time series data $\mathbf{S}$ can be denoted as a sequence $\mathbf{S} = [s_1, s_2, \ldots, s_T]$, where $T$ is the length of the time series, and $s_t$ is the observed value at time point $t$. For multivariate time series, each $s_t$ is a vector representing the observed values of all variables at time $t$.
The time series dataset $\mathcal{T}$ can be defined as:
$\mathcal{T} = \{(\mathbf{S}^1, c^1), (\mathbf{S}^2, c^2), \ldots, (\mathbf{S}^N, c^N)\}$
Here, each $(\mathbf{S}^i, c^i)$ represents the $i$-th time series sample and its corresponding class label.
The goal of the time series classification task is to learn a mapping function $f: \mathbf{S} \rightarrow c$ that can accurately predict the class label $c$ for 
a given time series $\mathbf{S}$. In other words, the task is to train a 
classifier $f$ such that for any new time series $\mathbf{S}^*$, the classifier can predict its belonging class $\hat{c}$
$\hat{c} = f(S^*)$
Where $\hat{c}$ is the predicted class by the classifier for the time series $\mathbf{S}^*$.

\subsection{Architecture of AdaFSNet}
The architecture of the AdaFSNet is shown in Fig. 2. It features a ResNet structure where the dropout layer is embedded in the OS-Block with a multi-kernel setup. Each kernel performs the same padding convolution with the input. We will individually introduce the various modules of AdaFSNet in the following content.
% \url{https://anonymous.4open.science/r/AdaFSNet-F7F7}

\subsubsection{OS-Block}
The Omni-Scale block (OS-Block) \cite{tang2020omni}, which serves as a component of 1D-CNN, automatically sets kernel sizes to efficiently cover receptive fields of all scales. It utilizes a set of prime numbers as kernel sizes to cover the best RF size across different datasets by transforming the time series through combinations of these prime-sized kernels. Particularly, for the kernel size configuration, we use $\mathbb{P}^{(i)}$ to denote the kernel size set of the $i$-th layers.   

\begin{equation}
    \mathbb{P}(i) = 
    \begin{cases} 
    \{p \in \mathbb{P} : p \leq p_k \}, & \text{if } i = 1, 2 \\
    \{2\}, & \text{if } i = 3
    \end{cases}
\end{equation}
where $p \in \mathbb{P} : p \leq p_k $ is a set of prime number from 1 to $p_k$. The value of $p_k$ is the smallest prime number that can cover all size of RF. So, we will have:
\begin{equation}
    \mathbb{S} = \{n \mid n = e \text{ or } n = e - 1, \text{ for some } e \in E\}
\end{equation}
The set $\mathbb{S}$ is derived from the combination of prime numbers in $\mathbb{P}(i)$, leading to a comprehensive set of even integers $\mathbb{E}$. This method ensures that $\mathbb{S}$ encompasses the full range of integer numbers since every real number must be either odd or even, and $\mathbb{E}$ is the subset of even numbers. With the right choice of the prime $p_k$, it's possible to cover any integer RF size within a given range. The utilization of Goldbach's conjecture allows for the assurance of encompassing all scales.

\subsubsection{Dropout module}
To address the multitude of options available to cover all Receptive Field (RF) sizes and minimize redundancy, the third layer of the network incorporates an attention module. This module selectively emphasizes channels with more efficient kernels, ensuring that the network utilizes the most effective features for processing while simultaneously reducing unnecessary complexity that could arise from redundant information.
At the end of the third layer, an attention mechanism processes its output, resulting in the creation of a channel attention map. This map is generated by leveraging the information from the preceding layer's output, applying the attention mechanism to enhance feature selection based on channel relevance.
Given the output of the previous convolutional layer as $\boldsymbol{\textit{U}} = [\boldsymbol{u}_1, \boldsymbol{u}_2, \ldots, \boldsymbol{u}_c] \in \mathbb{R}^{\textit{W}\times\textit{C}}$, where W is the width of the variate respectively, C is the number of channels. 

Initially, we determine the significance of each channel. This is achieved by converting the information from each variate into a channel-wise vector, utilizing global average pooling, a technique validated for its effectiveness by \cite{hu2018squeeze}. This method is crucial for assessing the relevance of different channels in the network.
Then, we determine the significance of each segment by averaging over each channel's features:
\begin{equation}
    \mathbf{v}_c = \frac{1}{\mathit{W}}\sum_{i=1}^{\mathit{W}}u_c(i)
\end{equation}
where $v_c$ is the significance value for the $c$-th channel. $c$-th element of $\boldsymbol{v}$.
To better capture channel-wise dependencies, the vector undergoes processing in a shared network, resulting in the creation of a channel attention map $\textbf{\textit{M}} \in \mathbb{R}^{1 \times C}$. This network comprises two fully connected (FC) layers paired with activation functions. The initial layer, aimed at reducing dimensionality, features parameters $\textbf{\textit{W}}_1 \in \mathbb{R}^{\frac{C}{r} \times C}$ and is followed by a ReLU function. Then, a dimension-increasing layer with parameters $\textbf{\textit{W}}_2 \in \mathbb{R}^{C \times \frac{C}{r}}$ leads up to a final Sigmoid function. This sequential arrangement is pivotal for enhancing the network's ability to discern channel-specific information.
These components are arranged in an alternating sequence, effectively managing the network's dimensional transformations.
Here, the reduction ratio 'r' is used to modify the bottleneck. This affects the map showing the relationships between channels. The operation, represented as $\textbf{\textit{F}}_{\textbf{\textit{v}} \rightarrow \textbf{\textit{M}}}$, defines these inter-channel connections, can be defined as:
\begin{equation}
    \textbf{\textit{M}} = \sigma(\textbf{\textit{W}}_2 \delta (\textbf{\textit{W}}_1 \textbf{\textit{v}}))
\end{equation}
where $\sigma$ and $\delta$ refer to the Sigmoid and ReLU, respectively.

After that, values in matrix $\textbf{\textit{M}}$ are sorted, and the top \textit{K} elements are selected as targets based on the drop probability $\gamma$. The channels corresponding to these top elements, identified as "1" in vector $\textbf{\textit{T}} \in \mathbb{R}^{1 \times \textit{C}}$, are designated as the target channels. The threshold value, identified as $\textit{M}_{topK}$ in $\textbf{\textit{M}}$, guides this selection process. This method effectively determines which channels are most crucial for the network's performance and can be defined as:
\begin{equation}
    K = \lceil \gamma C \rceil \quad T_p = \begin{cases} 
    1 & \text{if } M_p \geq M_{\text{topK}} \\
    0 & \text{otherwise} 
    \end{cases}   
\end{equation}
where $\textit{M}_p$ and $\textit{T}_p$ denote the $p$-th elements of $\textbf{\textit{M}}_p$ and $\textbf{\textit{T}}$.

For each segment corresponding to a target channel in a one-dimensional time series, we aim to pinpoint a subsequence with substantial discriminative information. Instead of using convolution operations with sizable kernels, which may lead to considerable computational overhead, we can leverage the continuity of time series data. By identifying the point with maximum value, we presume that adjacent top values within the sequence are likely to embody key features of the predominant signal pattern. Therefore, we select the point $a$ with the maximum value and discard a region of length $k$ centered around it. The boundaries of the target subsequence and the TargetDrop 
mask \( S = [s_1, s_2, \ldots, s_C] \in \mathbb{R}^{W \times C} \)
\begin{equation}
   w_1 = \max(a - \frac{k}{2}, 1) \quad w_2 = \min(a + \frac{k}{2}, W)
\end{equation}
This approach adapts the spatial attention mechanism to the temporal domain, selectively regularizing the time series data to enhance model generalization.

The final output for the time series data, after applying the TargetDrop mask, is denoted as \(\tilde{U} = [\tilde{u}_1, \tilde{u}_2, \ldots, \tilde{u}_C]\) and belongs to the space \(\mathbb{R}^{W \times C}\). The mask application and feature normalization process is as follows: 
\begin{equation}
    \tilde{u}_{z,n} = u_{z,n} \odot s_{z,n} \times \frac{\text{numel}(s_{z})}{\text{sum}(s_{z})}.    
\end{equation}
In this equation, \(\tilde{u}_{z,n}\) and \(s_{z,n}\) represent the \(n\)-th data point in the \(z\)-th channel of \(\tilde{U}\) and \(S\) respectively, \(\text{numel}(s_{z})\) is the count of time points in \(s_{z}\), \(\text{sum}(s_{z})\) is the count of unmasked time points, and \(\odot\) signifies the point-wise multiplication operation.
This method, adjusted for time series data, employs selective regularization across the data's temporal aspect to enhance model generalization and performance.

\subsubsection{Convolutional Block and Dense Connection}
Motivated by the impressive feature extraction abilities demonstrated by ResNet, we believe there is potential to further enhance ResNet's performance. This improvement can be achieved by integrating concepts from dense Convolutional Neural Networks (CNNs)\cite{huang2017densely}. The convolutional block is composed of eight fundamental 1D CNNs, each utilizing the ReLU activation function. Post-CNN feature extraction, these features undergo processing by a batch normalization (BN) layer. The convolutional process itself is carried out using a predetermined kernel size, which is preserved from the preceding dropout layer. This structure ensures efficient feature extraction and processing within the convolutional block. In this way, we can take full advantage of the OS-Block module's power to extract the most suitable RF size with the calculated convolution kernel. In addition, we introduce residual connection, which saves the output of os block module and adds it to the middle output of dense connection, so that it can learn more meaningful feature representation. Ultimately, we obtain the output class label by passing the data through both the global average pooling layer and the fully connected layer.

\section{Experiments}

\subsection{Benchmarks}

In our experiments, we selected the 34 datasets from the UCR and 21 datasets from the UEA Time Series Classification Repository \cite{bagnall2018uea}. The datasets are commonly used datasets of the baseline methods.

\begin{table}[ht]
    \centering
    \caption{INFORMATION ABOUT SOME UCR DATASETS.}
    \label{tab:datasets}
\resizebox{1\linewidth}{!}{
    \begin{tabular}{lcccc}
        \hline
        \textbf{Datasets} & \textbf{Train Size} & \textbf{Test Size} & \textbf{Length} & \textbf{Class} \\
        \hline
        Adiac           & 390  & 391  & 37 & 37 \\
        Beef            & 30   & 30   & 5  & 5 \\
        CBF             & 30   & 900  & 128 & 3  \\
        CricketX        & 390  & 390  & 300  & 12 \\
        Coffee          & 28   & 28   & 2  & 2 \\
        % DiatomSizeReduction & 16   & 306  & 4  & 25 \\
        ECGFiveDays     & 23  & 861  & 136  & 2  \\
        FaceAll         & 560  & 1960  & 131  & 14 \\
        % ItalyPowerDemand & 67   & 1029 & 2  & 24  \\
        FacesUCR        & 200 & 2050 & 131 & 14 \\
        FiftyWords      & 450 & 455 & 270 & 50 \\
        % Lightning7      & 70   & 73   & 7  & 319 \\
        GunPoint        & 50 & 150 &  150 & 2 \\
        OliveOil        & 30  & 30 & 570 & 4 \\ 
        Symbols         & 25   & 995  & 398  & 6 \\
        Trace           & 100  & 100  & 275  & 4 \\
        % TwoLeadECG      & 23   & 1139 & 2  & 82  \\
        \hline
    \end{tabular}
}
\end{table}

Table I lists information about these datasets. 
The UEA \& UCR Time Series Classification Repository divides each dataset into training and testing subsets. Consistent with common practices in TSC research, the training subset is utilized for developing TSC methods in our experiments, while the testing subset is employed to enhance classification accuracy. The accuracy of dataset $D$ is determined according to Eq.8, where $N_{test}$ represents the total count of test data points, and $N_c$ signifies the number of correctly classified points in the test dataset.

\begin{equation}
    \text{Accuracy} = \frac{N_c}{N_{test}}
\end{equation}

\subsection{Experiment Setup}

In line with the approach detailed in \cite{wang2017time}, our experiments adopt a learning rate of 0.001, a batch size of 16, and employ the Adam optimizer \cite{kingma2014adam}. Our methodology is implemented using PyTorch and the experiments are conducted on an Nvidia RTX 4090 GPU.

\subsection{Evaluation Metrics}
In all benchmark assessments, we follow the established conventions outlined in prior research as documented in \cite{8141873}. For evaluating the classification performance of models on UCR \& UEA time series datasets, we utilize the Mean Per-Class Error (MPCE). For a given model, the Per-Class Error (PCE) on the $k$-th dataset is calculated as follows:
\begin{equation}
PCE_k = \frac{e_k}{c_k}
\end{equation}
Here, \(e_k\) represents the classification error, and \(c_k\) is the count of categories in the $k$-th dataset. The overall MPCE for a model across all datasets is calculated as follows:
\begin{equation}
MPCE = \frac{1}{D} \sum_{k=1}^{D} PCE_k
\end{equation}
where \(D\) represents the aggregate number of datasets involved.
% For the multidimensional time series datasets, we used the accuracy to evaluate the classification performance of the models.
% \begin{equation}
% Acc = \frac{1}{M} \sum_{i=1}^{M} \mathbb{I}(f(x^{(i)}) = y^{(i)})
% \end{equation}
% where \(M\) is the number of samples in a dataset, \(\mathbb{I}(\cdot)\) is the indicator
% function, \(f(x^{(i)})\) represent the prediction of the \(i\)th sample, and \(y^{(i)}\) is
% its ground truth.

\begin{table*}[h!]
    \centering
    \caption{Comparing the error rate in TSC for six methods across 34 UCR datasets.}
    \begin{tabular}{lcccccccc}
        \hline
        Dataset & DTW & MCNN & InceptionTime & BOSS & ROCKET  & \textbf{AdaFSNet(ours)} & \textbf{AdaFSNet(FT)*} \\
        \hline
        \hline
        Adiac & 0.398 & 0.351 & 0.412 & \textbf{0.220} & 0.356 & 0.255 & 0.250 \\
        Beef & 0.366 & 0.433 & 0.366 & 0.233 & 0.266 & \textbf{0.033} & \textbf{0.033} \\
        CBF & 0.010 & 0.035 & 0.020 & \textbf{0} & 0.250 & \textbf{0} & 0.025 \\
        ChlorineCon & 0.385 & 0.300 & 0.455 & 0.340 & 0.195 & 0.180 & \textbf{0.155} \\
        CinCECGTorso & 0.480 & \textbf{0.078} & 0.090 & 0.125 & 0.160 & 0.210 & 0.180 \\
        Coffee & \textbf{0} & 0.047 & \textbf{0} & \textbf{0} & \textbf{0} & \textbf{0} & \textbf{0} \\
        CricketX & 0.326 & 0.456 & 0.212 & 0.255 & 0.265 & 0.200 & \textbf{0.180} \\
        CricketY & 0.256 & \textbf{0.185} & 0.221 & 0.208 & 0.195 & 0.215 & 0.195 \\
        CricketZ & 0.276 & 0.242 & 0.213 & 0.246 & 0.228  & 0.210 & \textbf{0.195} \\
        DiatomSizeR & 0.103 & 0.053 &\textbf{0.020} & 0.046 & 0.044 & 0.046 & 0.066 \\
        ECGFiveDays & 0.221 & \textbf{0} & \textbf{0} & \textbf{0} & \textbf{0} & \textbf{0} & \textbf{0} \\
        FaceAll & 0.235 & 0.305 & 0.210 & 0.245 & \textbf{0.185} & \textbf{0.185} & \textbf{0.185} \\
        FaceFour & 0.160 & 0.045 & \textbf{0} & 0.207 & \textbf{0} & \textbf{0} & \textbf{0} \\
        FacesUCR & 0.127 & 0.072 & 0.136 & \textbf{0.043} & 0.163 & 0.146 & 0.132 \\
        FiftyWords & 0.240 & 0.200 & 0.280 & 0.270 & 0.240 & 0.193 & \textbf{0.176} \\
        Fish & 0.177 & 0.087 & \textbf{0.021} & 0.121 & 0.056 & 0.041 & 0.033 \\
        GunPoint & 0.133 & \textbf{0} & \textbf{0} & \textbf{0} & \textbf{0} & 0.015 & \textbf{0} \\
        Haptics & 0.523 & 0.533 & 0.514 & 0.536 & 0.540 & 0.515 & \textbf{0.495} \\
        InlineSkate & 0.536 & 0.591 & \textbf{0.510} & 0.511 & 0.589 & 0.646 & 0.582 \\
        ItalyPower & 0.050 & 0.045 & 0.045 & 0.053 & 0.056 & 0.051 & \textbf{0.027} \\
        Lightning2 & \textbf{0.129} & 0.186 & 0.212 & 0.148 & 0.146 & 0.157 & 0.146 \\
        Mallat & 0.085 & 0.063 & 0.054 & 0.058 & 0.054 & 0.030 & \textbf{0.020} \\ 
        NonInvThoraxl & 0.210 & 0.066 & 0.212 & 0.161 & \textbf{0.068} & 0.085 & 0.079 \\
        OliveOil & 0.327 & 0.129 & 0.153 & 0.100 &\textbf{0.096} & 0.156 & 0.134 \\
        SonyAIBORobotl & 0.346 & 0.250 & \textbf{0.210} & 0.321 & 0.235 & 0.232 & 0.213 \\
        StarLightCurves & \textbf{0.010} & 0.120 & 0.126 & 0.021 & 0.039 & 0.073 & 0.043 \\
        SwedishLeaf & 0.218 & \textbf{0.028} & 0.141 & 0.072 & 0.152 & 0.042 & \textbf{0.028} \\
        Symbols & 0.050 & 0.038 & 0.031 & 0.032 & 0.174 & 0.078 & \textbf{0.022} \\
        Trace & \textbf{0} & \textbf{0} & \textbf{0} & \textbf{0} & \textbf{0} & \textbf{0} & \textbf{0} \\
        UWaveX & 0.301 & 0.201 & 0.250 & 0.241 & 0.243 & 0.222 & \textbf{0.178} \\
        UWaveY & 0.359 & 0.277 & 0.344 & 0.313 & 0.297 & 0.335 & \textbf{0.265} \\
        Wafer & \textbf{0.020} & 0.040 & \textbf{0.020} & 0.080 & 0.040 & 0.040  &  \textbf{0.020}\\
        wordSynonyms & 0.345 & \textbf{0.199} & 0.439 & 0.345 & 0.406 & 0.370 & 0.324 \\
        Yoga & 0.213 & \textbf{0.129} & 0.171 & 0.257 & 0.163 & 0.182 & 0.144 \\
        Win & 5 & 8 & 10 & 7 & 8 & 7 & \textbf{19} \\
        \hline
    \end{tabular}
\end{table*}

\subsection{Comparison with state-of-the-art methods}

Tables II and III provide a comprehensive comparison of our AdaFsNet with other Time Series Classification (TSC) models. Notably, our network consistently outperforms in both one-dimensional and multidimensional data scenarios. Other models, such as the Multi-scale Convolutional Neural Networks MCNN \cite{cui2016multi}, Fully Convolutional Networks (FCN) \cite{wang2017time}, ROCKET\cite{dempster2020rocket}, InceptionTime\cite{ismail2020inceptiontime} and BOSS\cite{schafer2015boss}, also exhibit strong performances. The MCNN model tackles the limitation of extracting features at a single scale by utilizing different rates of downsampling in time series data, thus allowing for multi-scale feature extraction. The FCN, often used as a benchmark, is enhanced into a ResNet-style classifier for time series by incorporating multiple FCN layers with residual connections. ROCKET achieves exceptional accuracy and speed by using random convolutional kernels to transform time series data, which are then used to train a linear classifier. InceptionTime, a highly effective ensemble model which employs multiple deep Convolutional Neural Networks. The BOSS model excels in identifying and filtering substructures in time series through Symbolic Fourier Approximation (SFA) and employs string matching for comparing series.

\begin{table}[htb!]
    \centering
    \caption{Evaluating the accuracy of five different methods across 23 datasets from the UEA datasets.}
\resizebox{1\linewidth}{!} 
{
    \begin{tabular}{lccccc}
        \hline
        Dataset & DTW & FCN & ResNet & MLP & \textbf{AdaFSNet(ours)} \\
        \hline
        Articulary & \textbf{0.987} & 0.953 & 0.853 & 0.353 & 0.973 \\
        AtrialFibrillation & 0.345 & 0.265 & 0.325& 0.400 & \textbf{0.525} \\
        BasicMotions & 0.975 & \textbf{1} & \textbf{1} & 0.275 & \textbf{1} \\
        CharacterTrajectories & \textbf{0.986} & 0.974 & 0.993 & 0.618 & 0.979 \\
        Cricket & \textbf{1} & 0.963 & 0.874 & 0.454 & \textbf{1} \\
        DuckDuckGeese & 0.544 & 0.572 & 0.581 & 0.456 & \textbf{0.622} \\
        Epilepsy & 0.965 & \textbf{1} & 0.961 & 0.273 & 0.985 \\
        ERing & 0.333 & 0.466 & \textbf{0.866} & 0.366 & 0.733 \\
        FaceDetection & 0.519 & 0.664 & 0.632 & 0.312 & \textbf{0.684} \\
        FingerMovements & 0.531 & 0.600 & 0.585 & 0.534 & \textbf{0.642} \\
        HandMovement & 0.271 & 0.227 & \textbf{0.405} & 0.383 & 0.383 \\
        Handwriting & 0.286 & 0.302 & 0.354 & 0.186 & \textbf{0.456} \\
        JapaneseVowels & 0.949 & 0.962 & 0.986 & 0.853 & \textbf{0.983} \\
        Libras & 0.872 & 0.872 & \textbf{0.944} & 0.325 & 0.901 \\
        MotorImagery & 0.558 & 0.510 & 0.510 & 0.258 & \textbf{0.534} \\
        NATOPS & 0.883 & \textbf{0.938} & 0.88 & 0.876 & 0.916 \\
        PenDigits & \textbf{0.975} & 0.938 & 0.962 & 0.952 & 0.962 \\
        PhoneTime & 0.172 & 0.218 & 0.232 & 0.097 & \textbf{0.215} \\
        RacketSports & 0.803 & \textbf{0.841} & 0.835 & 0.356 & 0.678 \\
        SelfRegulationSCP1 & 0.775 & 0.624 & 0.756 & 0.432 & \textbf{0.876} \\
        % SelfRegulationSCP2 & 0.539 & 0.50 & 0.554 & 0.138 & \textbf{0.588} \\
        % StandWalkJump & 0.2 & 0.33 & 0.33 & 0.120 & \textbf{0.61} \\
        UWaveGestureLibrary & 0.903 & 0.553 & 0.837 & 0.532 & \textbf{0.912} \\
        Win & 4 & 4 & 4 & 0 & \textbf{12} \\
        \hline
    \end{tabular}
}
\end{table}

Although other methods demonstrate strong performance in TSC, as evidenced in Tables II and III, the AdaFSNet model exceeds them in evaluation metrics, underlining its enhanced effectiveness.

To assess the proposed model performance, test results for both fine-tuned and non-fine-tuned versions of the model were presented, as shown in Table II. The non-fine-tuned results show improvements in accuracy for various datasets, like ChlorineCon, CricketX, CricketY, CricketZ, and FaceFour, with specific percentage gains noted for each. When fine-tuned, the AdaFSNet model not only achieves the highest average accuracy but also outperforms other models on most benchmark datasets, demonstrating its robust performance across a range of datasets.
Comparing results with and without fine-tuning the model, it's evident that fine-tuning significantly benefits certain datasets, like AREM, EEG, EEG2, Gesture Phase, HAR, HT Sensor, Ozone, and Action 3D, each showing notable percentage improvements. Other datasets exhibited consistent results regardless of fine-tuning. These benchmark tests also indicate that the AdaFSNet is versatile, applicable across various tasks and domains. 
In Table II, the comparison of classification error rates for 8 models on 34 datasets from the UCR collection is presented. The analysis reveals that conventional methods for time series classification, like DTW, BOSS, and MCNN, lead the charts in 5, 7, and 8 datasets respectively in terms of having the lowest error rates. This highlights their relative strengths in certain types of time series classification tasks.
Deep learning based models with good performance have emerged in recent years such as InceptionTime, ROCKET has 10 and 8 best results.

The data presented in Table III reveals that among five models evaluated across 23 UEA datasets, the proposed method outperforms others by achieving the highest classification accuracy on 12 datasets. This is a notable contrast to the performance of DTW, FCN, ResNet, and MLP models, which secured the top accuracy in 4, 4, 4, and 0 datasets respectively, which indicate that our method demonstrates enhanced accuracy across a wide range of UEA Time Series datasets.

\subsection{Baseline Comparison}
We compare our model with five baseline models across 85 UCR datasets as follows:
\begin{itemize}
    \item MLP: We utilize an MLP that consists of three fully-connected layers stacked sequentially, each containing 500 neurons. The ReLU function acts as the activation mechanism. Additionally, a dropout strategy is implemented with a rate of 0.8 to mitigate overfitting between these layers.
    \item CNN: The architecture includes two 1D convolutional layers that use the sigmoid function for activation, complemented by average pooling operations in the intervals between these layers.
    \item FCN: The configuration for the FCN is identical to that employed in the MLSTM-FCN as referenced in \cite{karim2019multivariate}.
    \item LSTM: A variety of LSTMs are trained, each differentiated by the count of hidden units, which range in the set {8, 16, 32, 64, 128}.
    \item ResNet: For the ResNet model, we adhere to the specifications outlined in the ResNet study found in \cite{ismail2019deep}.
\end{itemize}

In this section, AdaFSNet is trained without fine-tuning, using the same hyper-parameters as outlined in Section IV-B. To comprehensively evaluate our model within the constraints of page limits, Table IV lists the times of each method achieving the best performance. Among these, CNN, FCN, ResNet, and AdaFSNet are purely convolutional neural networks, while LSTM represents a different combination of neural network structures. The AdaFSNet, based on ResNet with added dense connections and specialized dropout modules, demonstrates significantly improved accuracy over baseline models on both datasets. This improvement is attributed to the full utilization of the OS-Block's potential to encompass all receptive fields, enhancing overall performance.

\begin{table}[htb!]
    \centering
    \caption{}
\resizebox{1\linewidth}{!} 
{
    \begin{tabular}{lcc}
        \hline
        Method & Baseline wins & \textbf{AdaFSNet(ours)} wins \\
        \hline
        MLP & 8 & \textbf{72}  \\
        CNN & 13 & \textbf{67}  \\
        FCN & 9 & \textbf{71}  \\
        LSTM & 22 & \textbf{58}  \\
        ResNet & 29 & \textbf{51}  \\
        LSTM-FCN & 34 & \textbf{46} \\
        \textbf{AdaFSNet(ours)} & - & -  \\
        \hline
    \end{tabular}
}
\end{table}

\subsection{Ablation Study}
We describe our evaluation of the effectiveness of the proposed model in terms of ablation analysis.

\subsubsection{Influence of the Dropout Module}
We measure performance by removing the dropout module and keep the 2Dense-Block on the UCR 85 \& UEA 30 datasets. We uniformly set the maximum epoch of training to 1500,  batch size of 16, and Adam\cite{kingma2014adam} optimizer. For each dataset, we test averaging three times. We present a quantitative comparison of results in Table V and list that. The numbers in Figure 4 indicate the number of times each baseline achieved the best results on each data set. We found that, when the data sets are more complex, the efficiency and accuracy of the model with the dropout layer removed decrease, and for some small data sets, the operating efficiency and accuracy do not change much.

\begin{table}[htb!]
    \centering
    \caption{}
\resizebox{1.0\linewidth}{!} 
{
    \begin{tabular}{lcc}
        \hline
         & UCR 85 archive \cite{bagnall2018uea} & UEA 30 archive\cite{bagnall2018uea}\\
        \hline
        OS-Block(norm) & 10 & 7  \\
        OS-Block+Dropout & 15 & 5  \\
        \textbf{AdaFSNet(ours)} & \textbf{55} & \textbf{18}  \\
        \hline
    \end{tabular}
}
\end{table}

\subsubsection{Influence of the Dense Connection}
To investigate the influence of the Dense Connection module, we conduct our experiment on the UCR 85 \& UEA 30 datasets.  The setting of hyperparameters is the same as above. Table VI shows the results, from which it can be seen that, with the resnet module, the accuracy of the model has been enhanced, and notably, for large and complex datasets, the increase in running speed is marginal. This efficiency is attributed to the extraction of the most suitable convolution kernel size at the dropout layer, which ensures high accuracy while not significantly adding to the computational burden.

\begin{table}[htb!]
    \centering
    \caption{}
\resizebox{1.0\linewidth}{!} 
{
    \begin{tabular}{lcc}
        \hline
         & UCR 85 archive \cite{bagnall2018uea} & UEA 30 archive\cite{bagnall2018uea}\\
        \hline
        OS-Block(norm) & 6 & 7  \\
        OS-Block+1Dense-Block & 12 & 22 \\
        OS-Block+2Dense-Block & 13 & 5  \\
        \textbf{AdaFSNet(ours)} & \textbf{49} & \textbf{18}  \\
        \hline
    \end{tabular}
}
\end{table}

% \begin{figure}[t]
% \begin{center}
%     % \includegraphics[scale=0.5, 
%     \includegraphics[width=1.0\linewidth, height=6.8cm]{figs/Figure_2.png}
% \end{center}
% \caption{Influence of the number of Dense-Blocks}
% \label{fig:teasor}
% \vspace{-0.6cm}
% \end{figure}
\begin{figure}[htbp]
\begin{center}
    \includegraphics[width=1.0\linewidth]{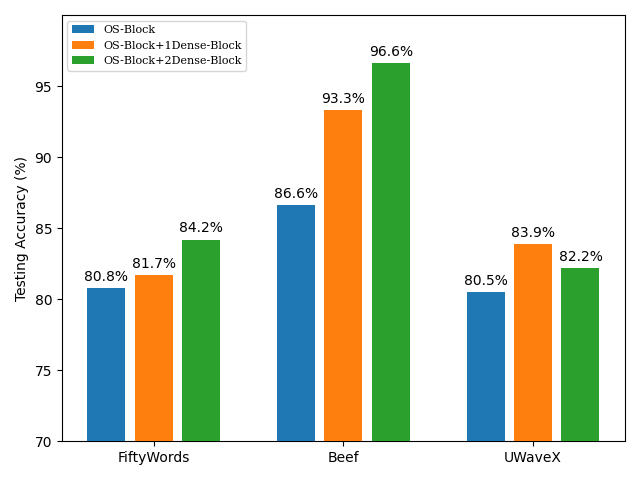}
\end{center}
\caption{Influence of the number of Dense-Blocks}
\label{fig:teasor}
\vspace{0.5cm}
\end{figure}

\begin{figure}[htbp]
\begin{center}
    \includegraphics[width=1.0\linewidth, height=4.5cm]{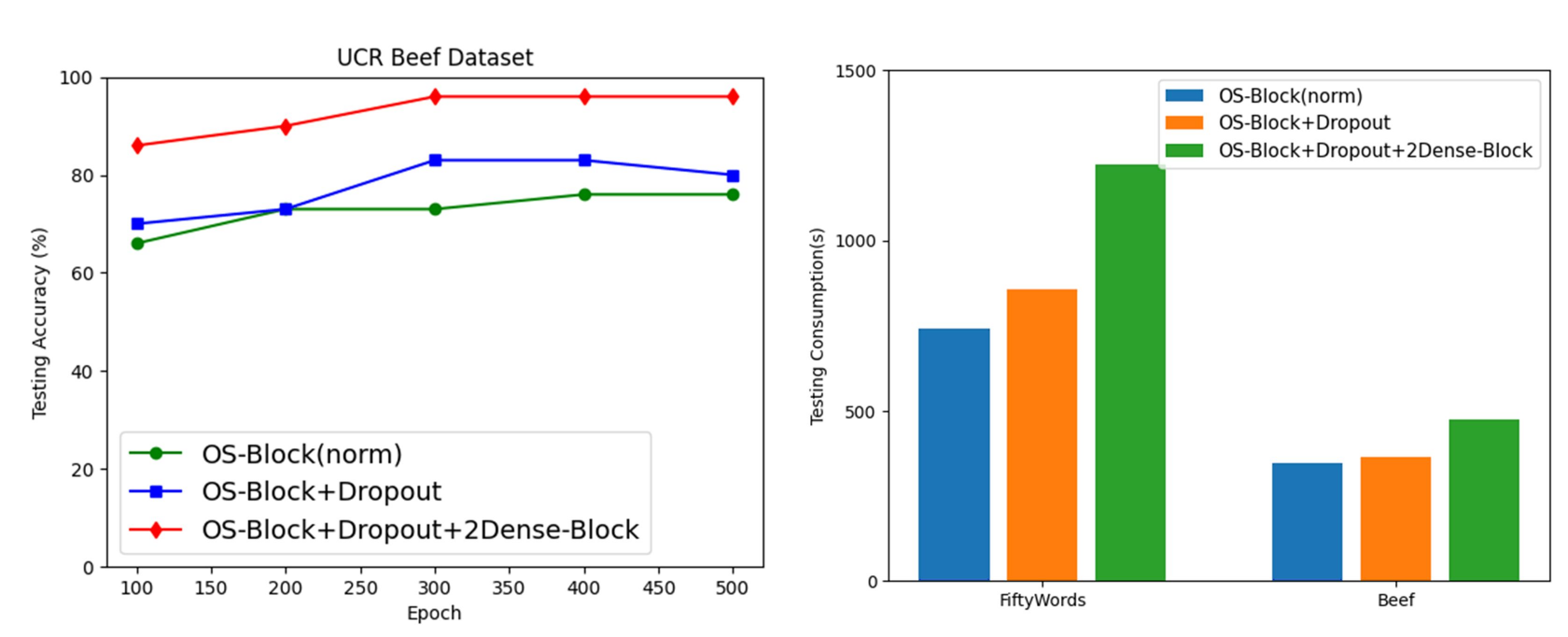}
\end{center}
\caption{Influence of the Dropout Module}
\label{fig:teasor}
\vspace{0.5cm}
\end{figure}

\section{Conclusion}
The paper presents AdaFSNet, which includes a ResNet framework integrating a dropout layer into the OS-Block, configured with a multi-kernel arrangement without
requiring adjustments in feature extraction scale. To maximize the efficiency of Receptive Field (RF) extraction in OS-Block, we incorporated a specialized dropout layer and two dense connections. Our proposed model presents an end-to-end framework that obviates the need for extensive preprocessing of the original dataset. Exhibiting robust generalization capabilities, it adaptively manages kernel size to obtain the most appropriate receptive field across one-dimensional and multi-dimensional time series data. We conduct experiments on the UCR \& UEA datasets and reveal that AdaFSNet is highly effective in capturing the optimal time scale in time series data. This robust scale capture capability of our model has enabled us to achieve a series of state-of-the-art results, outperforming the original OSCNN and multiple Time Series Classification (TSC) benchmarks.

\newpage

\bibliographystyle{IEEEtran}
\bibliography{IEEEabrv,reference}

\end{document}